\documentclass[a4paper, 10 pt, conference]{hsmr} 
\usepackage[utf8]{inputenc}
\IEEEoverridecommandlockouts                       
\usepackage{mathtools}
\usepackage{geometry}
\usepackage[labelsep=space]{caption}
\usepackage{newtxmath,newtxtext}
\usepackage{authblk}
\usepackage{pgfplots}
\usepackage{graphicx}
\usepackage[colorlinks,urlcolor=blue]{hyperref} 
\usepackage{newtxtext,newtxmath}
\usepackage{booktabs}

\pgfplotsset{compat=newest} 
 
\title{\LARGE \bf
S3-Tracker: Self-Supervised Surgical Tissue Tracking
With Contrastive Random Walks} 

\author{\LARGE Jiaming Zhang$^{1,*}$}
\author{\LARGE Zijian Wu$^{2,*}$} 
\author{\LARGE Mehran Armand$^{1,3}$}
\author{\LARGE Septimiu Salcudean$^{2}$} 

\affil{\Large\textit{$^{1}$Department of Computer Science, Johns Hopkins University,}\\ \Large\textit{$^{2}$Robotics and Control Laboratory, The University of British Columbia}\\\Large\textit{$^{3}$  The Institute for Integrative and Innovative Research, University of Arkansas} \\\Large\textit{$^{*}$ Equal contribution}\\ \Large\textit{jzhan282@jh.edu; zijianwu@ece.ubc.ca}}

\begin{document}

\maketitle
\thispagestyle{empty}
\pagestyle{empty}

\section*{INTRODUCTION}

Tracking surgical tissue in endoscopic videos is essential for computer-assisted intervention (CAI) and autonomous robotic surgery. Maintaining registration between real-time endoscopy and pre-operative imaging (CT/MRI) is particularly challenging due to soft tissue deformation. Robust point tracking enables continuous registration throughout surgery.

Recent Track-Any-Point (TAP) methods trained on large-scale real and synthetic datasets~\cite{karaevCoTracker32025} have demonstrated strong performance. However, dense ground-truth trajectories are challenging to obtain for surgical videos, given the presence of smoke, blood, specularities, and complex tissue deformation. Despite substantial efforts by the community, including novel methods~\cite{chen2025mfst, zhouEndoTTAP2025}, datasets~\cite{schmidtSurgical2024, cartucho2024surgt}, and open challenges~\cite{schmidt2025point}, robust surgical point tracking remains unsolved since supervised models are limited by the scale of reliably labelled dataset.

Motivated by these limitations, we consider using a self-supervised TAP approach to release the power of unlabeled surgical videos. We propose a method that establishes global pixel correspondences and infers point trajectories using contrastive random walks (CRW). Trained without annotations, our method achieves a promising performance comparable to existing semi-supervised approaches, and it's able to handle tissue deformations implicitly, demonstrating the feasibility of self-supervised point tracking in surgical environments.

\section*{MATERIALS AND METHODS}

\begin{figure}[t]
\centering
\includegraphics[width=\columnwidth]{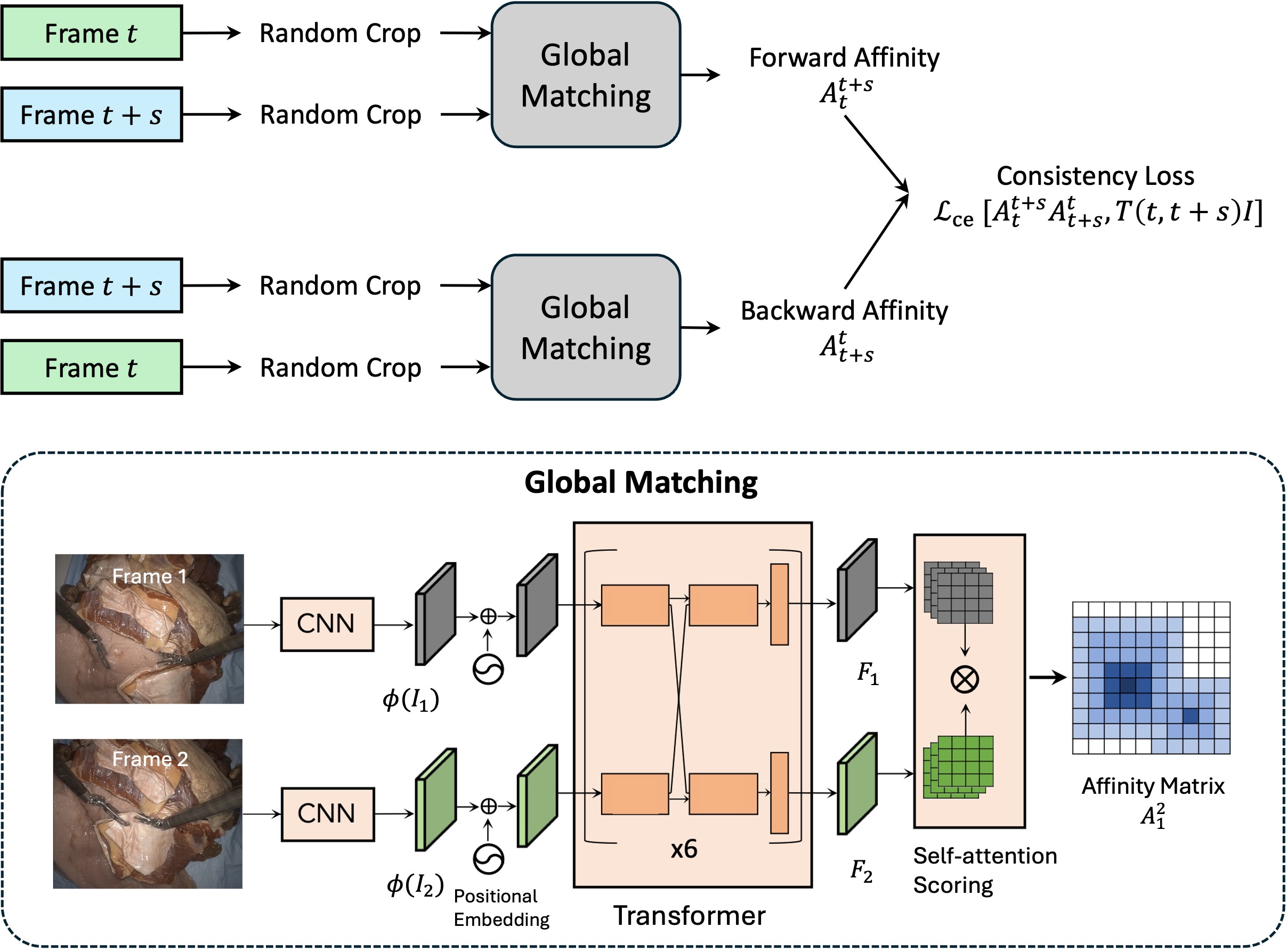}
\caption{ Model architecture. For any successive frames, we extract the feature vectors using a ResNet backbone and perform self- and cross-attention over the feature vectors to get the affinity matrix. We calculate the affinity matrices in both forward and backward directions and minimize the consistency loss between them.}
\label{fig_overview}
\end{figure}

Motivated by GMRW \cite{shrivastavaSelfsupervised2025}, our model learns long-range point correspondences through cycle-consistent contrastive random walks on a space--time graph, without requiring ground-truth trajectories. Given a video $ V = \{ I_t \}_{t=1}^{T}$  whose frames have $H \times W$ pixels, i.e. $ \quad I_t \in \mathbb{R}^{H \times W \times 3} $. The goal is to estimate trajectories $\mathbf{p}^k_t = (x^k_t, y^k_t)$ and visibility indicators $v^k_t \in \{0,1\}$ for arbitrary query points $q_k = (t_k, x_k, y_k)$. For every $I_t$, visual features are extracted using a CNN backbone and augmented with 2D positional encodings.

\subsubsection{Global Matching Architecture}
For a consecutive frame pair $(I_t, I_{t+1})$, stacked self-attention and cross-attention layers produce correlation features $F_t$ and $F_{t+1}$. The random-walk transition matrix is expressed as the cross attention between two feature vectors:
\[
A^{t+1}_t = \mathrm{softmax}\!\left( \frac{F_t F_{t+1}^{\top}}{\sqrt{d}} \right)
\]
where $A^{t+1}_t$, the affinity matrix, represents the probability of matching any spatial location in frame $t$ to frame $t+1$. And $d$ is the dimension of the feature vectors.

\subsubsection{Self-Supervised Learning via Contrastive Random Walks}
Training is performed using a CRW objective on palindrome sequences $I_t$ $\rightarrow$ $I_{t+1}$ $\rightarrow $ $I_t$. The model is supervised by enforcing that the composed transition returns to the starting location. Therefore, the contrastive random walk loss $\mathcal{L}_{\mathrm{CRW}}$ is defined as: 
\[
\mathcal{L}_{\mathrm{CRW}} 
= \mathcal{L}_{\mathrm{CE}}\!\left( A^{t}_{t+1} A^{t+1}_{t}, \; \mathbb{I} \right),
\]
where $\mathcal{L}_{\mathrm{CE}}$ denotes cross-entropy loss and $\mathbb{I}$ is the identity matrix. To avoid the shortcut solution often found in CRW, we apply different affine transformations $T_f$, $T_b$ to the forward cycle and backward cycle, respectively.

\subsubsection{Inference}
At inference, visibility is determined using a ResNet-based classifier and a cycle-consistency check: points are marked invisible when forward-backward displacement exceeds a threshold. To reduce drift, tracking rolls back to the last visible frame when invisibility is detected.

\section*{RESULTS}
We used an Nvidia A100 40G GPU to train and evaluate the model. Models are trained on STIR~\cite{schmidtSurgical2024} and SurgT~\cite{cartucho2024surgt} without using sparse point labels. Points are initialized in the first frame and tracked in a streaming manner to the last frame. Performance is evaluated using End Point Error (EPE) between predicted points $\hat{p}$ and ground truth $p$. Accuracy under threshold $T_i$ is defined as:
\[
\delta^{T_i} = \frac{1}{N}\sum_{n=1}^{N} f(\left| \hat{p} - p \right|, T_i)
\]
where $T$ = [4, 8, 16, 32, 64] px for full 1024 × 1280 images. N is the total number of points being tracked.$\left| \hat{p}-p \right|$ is the Euclidean EPE. $f$ is the indicator function, used to count the number of points under the distance threshold. The accuracy metric $\delta^{avg} = \frac{1}{I} \sum_{i=1}^I \delta^{T_i} $ is averaged across all thresholds with same weights. Here $I$ is the number of thresholds used. Tab.~\ref{tab:quant_results} compares state-of-the-art methods on STIR, which provides ground truth at the first and last frame of each video. With explicit visibility modeling, our method achieves performance comparable to Multi-Flow Tracker (MFT) while reducing inference latency. Although supervised models yield lower EPE, they operate offline and do not support frame-to-frame streaming inference.

\begin{table}[h]
\centering
\caption{Quantitative comparison of semi-supervised and unsupervised methods across four evaluation metrics.}
\label{tab:quant_results}
\begin{tabular}{lccc}
\toprule
\textbf{Method} &  $\delta^{avg}$ $\uparrow$ & EPE (px) $\downarrow$ & Latency (ms) $\downarrow$\\
\midrule
\multicolumn{4}{l}{\textit{Semi-supervised Methods}} \\
\midrule
Endo-TTAP \cite{zhouEndoTTAP2025} & 0.784 & 13.32 & -- \\
CoTracker  & 0.611 & 34.66 & -- \\
CoTracker3 \cite{karaevCoTracker32025} & 0.753 & 16.80 & -- \\
\midrule\midrule
\multicolumn{4}{l}{\textit{Unsupervised Methods}} \\
\midrule
MFT(2024)\cite{schmidt2025point}  & 0.776 & 18.32 & 417.83 \\
Ours (w/o Visibility)  & 0.684 & 26.13 & 529.47 \\
Ours & 0.708 & 22.65 & 301.47 \\

\bottomrule
\end{tabular}
\end{table}

\begin{figure}[htp]
\centering
\includegraphics[width=0.95\columnwidth]{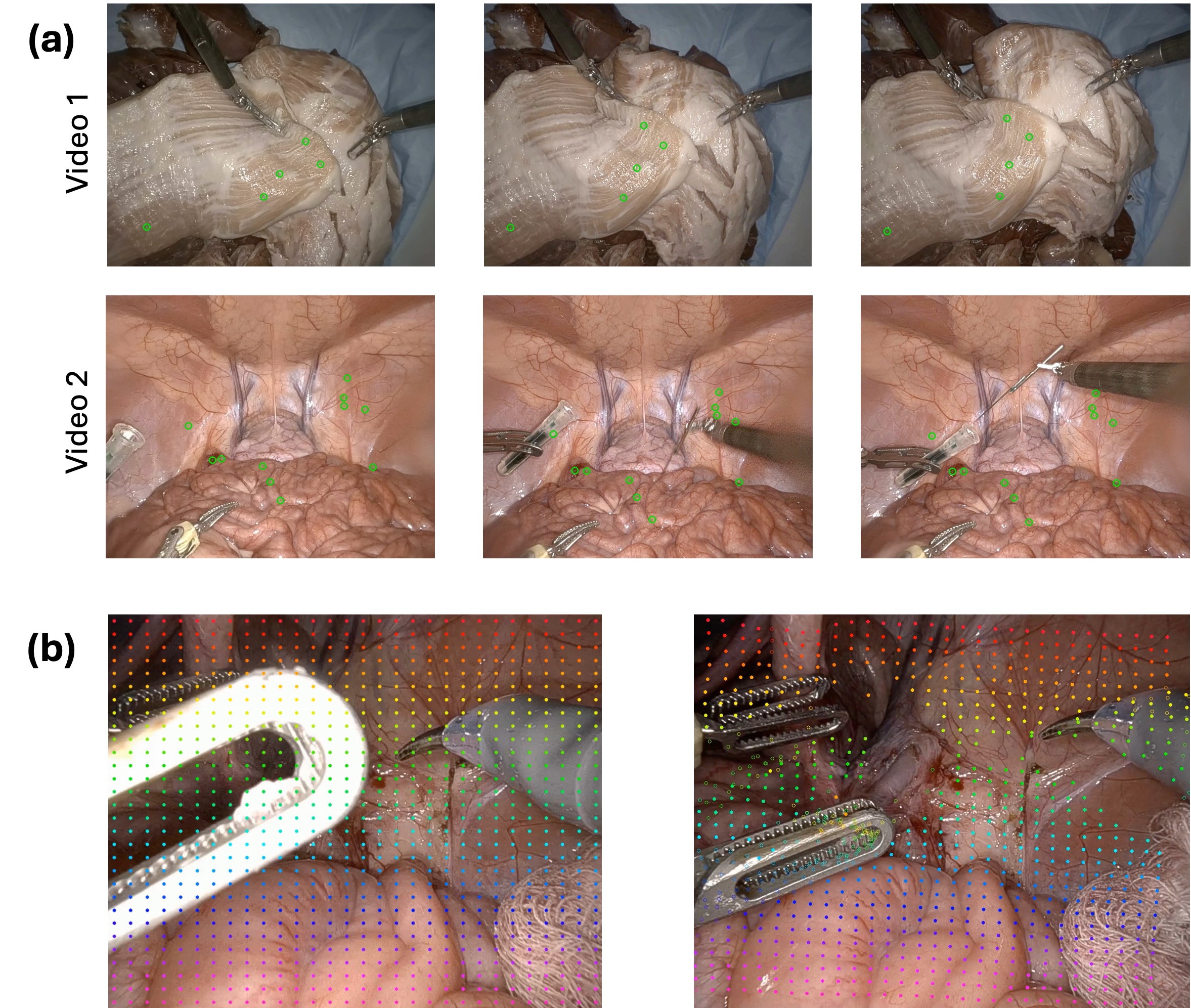}
\caption{(a) Point tracking results on challenging videos. The tracked points are marked as green circles. (b) The result of the visibility module determines the visibility of a large number of points. 30 $\times$ 30 points are initially sampled and are tracked by CoTracker3.}
\label{fig_result}
\end{figure}

Fig.~\ref{fig_result} illustrates qualitative results, including robust tracking under low texture similarity and long-term occlusions. In (a), our tracker can successfully track the points even though they are selected on a region where the textures are similar (video 1) and when there are occlusions (video 2). 

\section*{DISCUSSION}

This work shows that self-supervised point tracking is a viable alternative to supervised approaches for surgical videos, where dense annotations are hard to obtain. Despite the absence of supervision, the model achieves competitive accuracy with semi-supervised methods through global correspondence learning and contrastive random walks. Importantly, the proposed visibility module reduces inference latency, enabling efficient frame-to-frame tracking at 3 Hz and making the approach more suitable for near-real-time clinical deployment.

\nocite{*}
\bibliographystyle{IEEEtran}
\bibliography{TrackAnyPoint}

\end{document}